\pdfoutput=1
\documentclass[letterpaper, 10 pt, conference]{ieeeconf}  %

\IEEEoverridecommandlockouts                              %

\makeatletter
\let\NAT@parse\undefined
\makeatother
\usepackage[square,numbers,sort&compress]{natbib}

\usepackage{amsmath}
\usepackage{amssymb}

\usepackage[table,dvipsnames]{xcolor}
\usepackage{bm}
\usepackage{graphicx}
\usepackage{float}
\usepackage{subfig}
\usepackage{amsmath}
\usepackage{amssymb}
\usepackage{booktabs}
\usepackage{algorithm}
\usepackage{algpseudocode}
\usepackage{bbm}

\newcommand{\para}[1]{\medskip\noindent\textbf{#1. }}
\newcommand{\paras}[1]{\smallskip\noindent\textbf{#1. }}

\newcommand{\expect}{\mathbb{E}}
\newcommand{\prob}{P}
\newcommand{\reals}{\mathbb{R}}

\newcommand{\obs}{o}
\newcommand{\img}{I}
\newcommand{\proprio}{q}
\newcommand{\imgSpace}{\mathcal{I}}
\newcommand{\proprioSpace}{\mathcal{Q}}

\newcommand{\lang}{\ell}
\newcommand{\langSpace}{\mathcal{L}}

\newcommand{\policy}{\pi} %
\newcommand{\action}{a}
\newcommand{\acttraj}{\mathbf{\action}}
\newcommand{\actionSpace}{\mathcal{A}}

\newcommand{\activ}{z}

\newcommand{\func}{f}
\newcommand{\feat}{\zeta}

\newcommand{\probe}{\phi}

\newcommand{\traj}{\bm{\tau}}
\newcommand{\success}{Y}
\newcommand{\horizon}{H}

\newcommand{\score}{s}
\newcommand{\FPR}{\operatorname{FPR}}

\newcommand{\DT}{\operatorname{DetectTime}}

\newcommand{\progress}{\feat}

\newcommand{\thresh}{\delta}

\newcommand{\ours}{P^2}

\definecolor{monitorBlue}{HTML}{0048A6}
\definecolor{monitorRed}{HTML}{A8324A}
\definecolor{monitorTeal}{HTML}{0D948F}
\definecolor{monitorLightBlue}{HTML}{0091FF}
\definecolor{monitorPurple}{HTML}{740CAD}
\definecolor{monitorOrange}{HTML}{FF9500}
\definecolor{monitorPurpleLight}{HTML}{CDB3FF}
\newcommand{\methodFailInput}{\textcolor{monitorBlue}{FAIL-Detect (lopO)}}
\newcommand{\methodRndOE}{\textcolor{monitorBlue}{FIPER (RND-OE)}}
\newcommand{\methodACE}{\textcolor{monitorRed}{FIPER (ACE)}}
\newcommand{\methodSTAC}{\textcolor{monitorRed}{STAC}}
\newcommand{\methodSafeMLP}{\textcolor{monitorTeal}{SAFE-MLP}}
\newcommand{\methodSafeLSTM}{\textcolor{monitorTeal}{SAFE-LSTM}}
\newcommand{\methodFailFeature}{\textcolor{monitorLightBlue}{FAIL-Detect (logpZO)}}

\newcommand{\methodRobometerProgress}{\textcolor{monitorPurple}{Robometer (progress)}}
\newcommand{\methodRobometerSuccess}{\textcolor{monitorPurple}{Robometer (success)}}
\newcommand{\methodOurs}{\textcolor{monitorOrange}{Progress Probe \ensuremath{\boldsymbol{(\ours)}}}}
\newcommand{\methodRobometerFTProgress}{\textcolor{monitorPurpleLight}{Robometer FT (progress)}}
\newcommand{\methodRobometerFTSuccess}{\textcolor{monitorPurpleLight}{Robometer FT (success)}}

\usepackage[hidelinks]{hyperref}

\title{\LARGE \bf Progress Probes $P^2$: Lightweight Runtime Monitoring for Vision-Language-Action Models}

\newif\ifhtmlauthors
\makeatletter
\@ifundefined{lx@add@authors}{\htmlauthorsfalse}{\htmlauthorstrue}
\makeatother
\ifhtmlauthors
\author{Atiksh Bhardwaj \\ Carnegie Mellon University \\ Cornell University
\and Edward W. Duan \\ Cornell University
\and Junwon Seo \\ Carnegie Mellon University
\and Prithwish Dan \\ Cornell University
\and Wei-Chiu Ma \\ Cornell University
\and Andrea Bajcsy \\ Carnegie Mellon University \\ Equal advising
\and Preston Culbertson \\ Cornell University \\ Equal advising}
\else
\author{Atiksh Bhardwaj$^{1,2}$, Edward W. Duan$^{2}$, Junwon Seo$^{1}$, Prithwish Dan$^{2}$,\\
Wei-Chiu Ma$^{2}$, Andrea Bajcsy$^{1,\dagger}$, Preston Culbertson$^{2,\dagger}$%
\thanks{$^{1}$Carnegie Mellon University. $^{2}$Cornell University. $^{\dagger}$Equal advising.}%
}
\fi

\begin{document}
\maketitle
\thispagestyle{empty}
\pagestyle{empty}

\begin{abstract}
    Vision-Language-Action (VLA) models can perform diverse manipulation tasks directly from visual observations and natural language instructions, but they remain prone to mistakes and unpredictable failures. Deploying these policies without continuous human supervision will require robust runtime monitors that detect when the policy has gone off-course. Existing monitors use external foundation models, such as vision-language models (VLMs) to supervise the policy, introducing significant complexity and computational cost for monitoring at runtime. In this work, we investigate an alternate approach: \textit{can a VLA's own internal activations provide the information needed to monitor its execution?} Our method trains linear probes on policy activations from demonstrations to estimate progress, then flags executions that deviate from expected progress. In experiments with the open-source VLA $\pi_{0.5}$ and across simulated and real-world manipulation tasks, this approach is competitive with or outperforms VLM-based monitors in task failure detection, as well as methods which require failure data for training. These findings suggest that simple probes of VLA internals can be used to develop reliable runtime monitors, opening a path toward building policies that can act reliably and monitor their execution without external foundation models.

\end{abstract}

\section{Introduction}
\label{sec:introduction}

Vision-Language-Action models (VLAs) have made strong progress toward the goal of generalist, language-conditioned robot policies. Models such as $\pi_{0.5}$~\citep{intelligence2025pi05} and \texttt{GR00T}~\citep{nvidia2025gr00t} can execute diverse, multi-stage manipulation tasks across a variety of tasks and embodiment and, especially after task-specific fine-tuning, can achieve relatively high mutli-task success rates. However, even well-trained policies still experience a wide variety of failure modes, (e.g., poor language grounding \cite{kwok2026cover}, deadlocks, and mistakes \cite{gu2025safe}) which are often sudden and difficult to predict.

A natural framework for addressing such failure modes during deployment is \emph{runtime monitoring:} using an external verifier or monitor to detect when the policy's execution is straying from typical/successful deployment.
Existing approaches obtain this monitoring signal from several sources. Sentinel~\citep{agia2024sentinel} uses an external VLM, along with an action consistency heuristic, to judge when the policy is failing to make progress on a particular task. Generalist reward models, such as Robometer~\citep{liang2026robometer}, learn to judge task progress from datasets of robot videos. Other methods instead use policy internals directly for failure detection. For example, SAFE~\citep{gu2025safe} trains a failure classifier to predict rollout success directly from policy activations, while FIPER~\citep{romer2025failure} combines an internal-based anomaly score and an uncertainty metric over actions to flag anomalous rollouts.

In this work, we investigate a different use of policy internals for runtime monitoring.
\begin{figure*}[!tb]
    \centering
    \includegraphics[width=\textwidth]{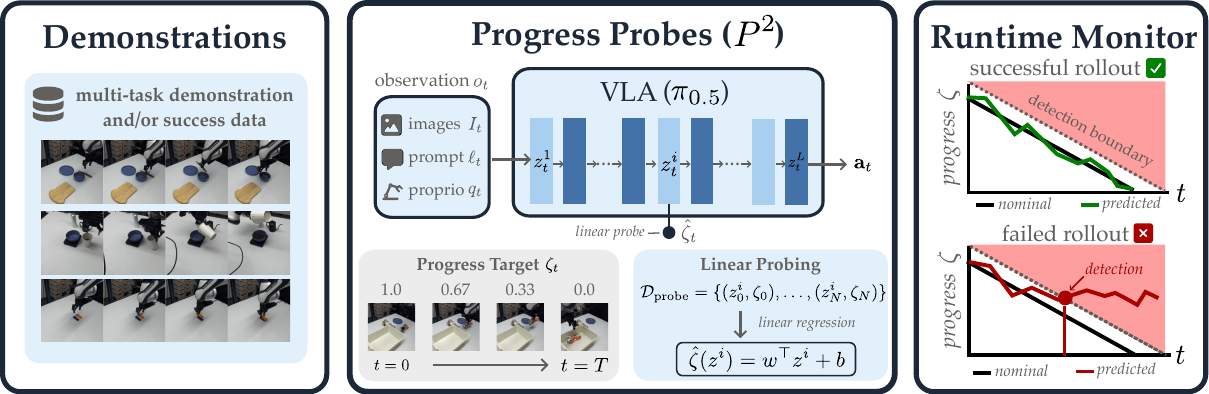}
    \caption{Progress Probes ($\ours$) decode task progress directly from the VLA’s internal activations. Using only successful demonstrations, we assign each timestep a normalized progress  $\feat_t$ and fit a linear probe to predict it from \(z_t^i\), the activation at layer \(i\). At runtime, successful executions follow the expected progression through the task, whereas failed executions cause the decoded progress to lag or deviate from this nominal trajectory. Thus, the policy’s own representations enable us to construct a lightweight runtime monitor without failure data or an external foundation model.}

    \vspace{-1.5em}
    \label{fig:figure_1}
\end{figure*}
Specifically, we build on recent successes in \textit{mechanistic interpretability} for VLAs~\citep{buurmeijer2026observing, lu2025probing}, which demonstrated that the activations of the VLM backbone contain easily-decodable representations of task-relevant features such as end-effector pose, object states, and task phase.
We hypothesize that a VLA's internal activations may also encode \textit{task progress}: how far the policy has progressed through a task. In successful executions, progress advances predictably over time (monotonically decreasing), while in suboptimal or failed executions, the progress probe falls behind the elapsed time. Furthermore, task progress can be learned entirely from successful executions and can be labeled without human supervision.
\begin{quote}
    \centering
    \emph{We propose progress probes---linear decodings from the VLA's activations to task progress---as lightweight internal-based runtime monitors.}
\end{quote}

We study progress probes by training them on the $\pi_{0.5}$ \cite{intelligence2025pi05} VLA
using a dataset of success-only task executions. At deployment, runtime monitoring consists of taking the VLA's internal activations (that are already computed as part of the forward pass of inference), passing them through the linear progress probe, and measuring the deviation between the current estimated task progress and the expected task progress computed from successful demonstrations. Not only does our probe-based monitor require only successful data for training, but it also runs between $5\times$ and $20,000\times$ faster than an external foundation-model monitor.

Across simulated and real-world manipulation tasks, we find that linear probes can indeed accurately predict task progress for demonstrations, and that they provide an effective signal for detecting unsuccessful policy rollouts. Our monitor achieves competitive or superior failure detection performance with both Robometer and SAFE baselines, while providing fast inference and requiring only successful demonstrations for training. This result indicates that even when VLAs fail to generate successful actions, their activations still retain information useful for detecting task progress and failure, which can be obtained using a simple linear readout. Specifically, our contributions are as follows:

\begin{enumerate}
\item We demonstrate empirically that linear probes trained over demonstration data can accurately decode task progress from $\pi_{0.5}'s$ vision-language representations, across simulated and real-world manipulation tasks.
\item We develop a minimalist progress-based runtime monitor which uses these probes to detect deviations from expected progress at runtime, without failure data or an external VLM.
\item We evaluate our monitor on 6 VLABench tasks and 6 real-world manipulation tasks, achieving AUROC scores of 0.788 and 0.78 respectively, matching or exceeding competitive baselines including SAFE and Robometer.
\end{enumerate}

\section{Related Work}
\label{sec:relatedworks}
\para{Interpretability of Robot Policies}
Mechanistic interpretability seeks to understand the behavior of neural networks by studying their internal representations and computations.
Over the past decade, researchers in computer vision and language modeling have developed tools to understand what neurons learn and how semantic concepts and structured knowledge are encoded and used.
One particularly exciting perspective is the \emph{linear representation hypothesis}~\citep{mikolov2013linguistic, park2024linear}, which posits that high-level concepts are encoded as directions in activation space and that interventions along concept directions can steer model behavior~\citep{zou2023representation}.
This allows one to connect the study of representations to their functional role and reveal behaviors not visible in model output \cite{hubinger2024sleeper, macdiarmid2024simple}.

More recently, researchers have extended similar ideas to robotic VLAs, investigating how their internal representations encode semantic concepts~\citep{haon2025mechanistic}, motion primitives~\citep{swann2026sparse}, and physical and relational states~\citep{buurmeijer2026observing, lu2025probing}.
Our work contributes to this growing body of interpretability research on robotic VLAs by investigating whether linear probes trained only on successful demonstrations can extract a \emph{task progress} signal from internal activations.
Beyond decoding, we further show that this signal can be turned into a lightweight runtime monitor that detects departures from expected progress, demonstrating how VLA internals can support failure detection without an external foundation model.

\para{Runtime Monitoring of Robot Foundation Models}
Learned robot policies, including VLAs, remain prone to unexpected failures during deployment, motivating techniques for monitoring policies at runtime and detecting failures as they occur. Existing runtime monitors differ primarily in the signals they use to recognize such failures. One approach is to use deviations from successful behavior, flagging observations or policy outputs that appear anomalous relative to successful executions \cite{xu2025can, romer2025failure}. Another is to leverage an external evaluator, such as a VLM judge \cite{agia2024sentinel, kwok2025robomonkey, wu2025foresight} or a learned failure predictor  \cite{gu2025safe} trained directly over policy internals. More general reward models can also serve as runtime monitors by recognizing abnormal task progression \cite{liang2026robometer}. Overall, several design decisions remain open for runtime monitoring, including whether monitoring requires an external validator, whether it must be trained on failure data, and whether the policy's own representations provide useful signals for monitoring. We study the latter hypothesis by probing these representations for task progress and detecting failures when estimated progress deviates from the progression expected during successful execution.

\section{Background: Mechanistic Interpretability \\ via Linear Probing}
\label{sec:bg}
We focus on a class of mechanistic interpretability techniques called \textit{probing} \cite{belinkov2022probing} and provide a brief overview.

\para{Setup} Let the VLA policy be denoted by $\policy(\acttraj_t \mid \obs)$ where $\acttraj_t = (\action_t, \hdots, \action_{t+h}) \in \actionSpace^h$ is an $h$-step action chunk and the input observation $\obs_t = (\img_t, \proprio_t, \lang_t)$ consists of an image $\img \in \imgSpace$, proprioceptive data $\proprio \in \proprioSpace$, and a language prompt $\lang \in \langSpace$.
Many VLA policies, including the one we study $\pi_{0.5}$~\citep{intelligence2025pi05}, are built on top of vision-language backbones (here, PaliGemma~\citep{beyer2024paligemma}) and use the transformer architecture; thus, we focus on this model class throughout the paper.

\para{Activation Extraction \& Probe Dataset Construction} The input observation sequence $\obs$ is first tokenized and then processed via a sequence of $L$ layers of transformer blocks in $\policy$, producing a sequence of intermediate activations at each layer.
Let the activation for any layer $i$ be denoted by $\activ^i \in \mathbb{R}^d$.
Each activation is computed sequentially via a residual function $\func^i: \mathbb{R}^d \to \mathbb{R}^d$,\begin{equation*}
    \activ^{i + 1} = \activ^i + \func^{i} (\activ^i), \quad i \in \{0, \dots, L-1\}.
\end{equation*}
Given a dataset of $N$ robot trajectories, $\mathcal{D} = \{(\obs_t, \action_t)\}_{n=1}^{N}$ where $t = 0, \dots, T$, we extract\footnote{Each activation $\activ_t^i$ has shape $d \times k$, where $d$ is the size of the activation and $k$ is the number of tokens; we mean-pool across the tokens to get $d$.} the activations from the VLA of interest by feeding in each observation $\obs_t$ and computing the intermediate layer activations $(\activ_t^0, \activ_t^1, \dots, \activ_t^L)$.
Let $\feat$ be the target feature we seek to linearly ``probe out'' of the VLA. For each trajectory and timestep in $\mathcal{D}$, we label the target feature.
Thus, we finally create our probe training dataset which consists of internal activations and the target feature labels for each of the $N$ trajectories:
$$\mathcal{D}_\text{probe} = \{(\activ_t^{0,n}, \activ_t^{1,n}, \dots, \activ_t^{L,n}, \feat^n_t)\}_{n=1}^N.$$

\para{Linear Probing} Our goal is to fit a linear function on the extracted activations.
Without loss of generality, consider a layer $i$ and a scalar target feature $\zeta \in \mathbb{R}$.
Our \textit{linear probe} \cite{alain2017understanding} is a map from the activations to the target feature:
$$\probe(\activ^i) = w^\top \activ^i + b,$$
with parameters $w, b$, trained on the train data split in $\mathcal{D}_\text{probe}$.

\section{Problem Formulation: Runtime Monitoring}
\label{sec:problem}
Our goal is to detect, during execution, when a VLA policy $\policy(\acttraj_t \mid \obs_t)$ is unlikely to complete its task. We follow the finite-horizon runtime monitoring setup of Sentinel~\citep{agia2024sentinel}.

\para{Setup} We observe a stream of robot observations and actions up until any time $t \geq 0$: let $\traj_t = (\obs_0, \action_0, \ldots, \obs_t)$ be this observation-action trajectory. Let $\horizon$ denote the maximum task horizon.
Finally, we define the rollout outcome $Y \in \{0,1\}$, where $\success=1$ if the policy reaches a successful state within $\horizon$ steps and $\success=0$ otherwise.

\para{Monitor} A \textit{runtime monitor}, $\score(\traj_t) \in \reals,$ is a map from the observation-action trajectory to a scalar value measuring how much the data differs from successful or ``nominal'' data (i.e. larger values imply more different from successful).
Given a threshold $\thresh \in \mathbb{R}$ and a trajectory $\traj$, let the \textit{detection time} (i.e., the time at which the monitor detects deviation from success) be $T_\thresh = \inf\{t \mid \score(\traj_t) > \thresh \}$, with $T_\thresh = \infty$ if the monitor never fires.

\para{Monitor Design Objective} The objective of runtime monitoring is to design a $\score(\traj)$ which (a) reliably discriminates successful trajectories from failures and (b) detects failures early. Formally, this is a multi-objective design problem:
\begin{equation}
    \min_\score \expect_{\traj} \big[\FPR(\delta, \traj; \score) + \lambda \cdot \DT(\delta, \traj; \score) \big],
    \label{eq:monitor-objective}
\end{equation}
for a given threshold $\thresh$, where $\lambda > 0$ is a parameter balancing the two objectives.
Here, the first term is the false positive rate of the monitor, $\FPR(\thresh, \traj; \score) = \prob(T_\thresh \leq H \mid Y = 1)$. The second term is the mean detection time on failed trajectories, $\DT(\thresh, \traj; \score) = \expect[T_\thresh \mid \success=0, T_\thresh \leq \horizon]$. An ideal runtime monitor minimizes false positives while also firing early under the true trajectory distribution (i.e., $\tau$ should ideally be drawn from the distribution of closed-loop evaluation trajectories).

\para{Goal} We hypothesize that we can design $\score(\traj)$ to be a linear probe (described in Sect.~\ref{sec:bg}) trained on the VLA's activations, achieving monitoring quality while being lightweight to run at inference time.

\section{Progress Probes as Runtime Monitors}
\label{sec:methods}

In this work, we propose using \textit{progress probes} ($\ours$), lightweight linear readouts of task progress from a VLA's internal activations, as runtime monitors. Our key insight is that a VLA's internal activations encode how far the policy has progressed through a task, and in successful executions this encoded progress advances predictably (monotonically) with elapsed time. Suboptimal or failed behaviors violate this relationship, and thus the progress probe falls behind the elapsed time.
Practically, task progress is also a useful signal for designing a runtime monitor because it can be easily labeled without human supervision from offline data (because it decreases monotonically on successful data) and can be learned without explicit failure data.

\para{Progress Probes} Following Sec.~\ref{sec:bg}, we train linear probes to predict normalized task progress from the policy's internal activations, given a dataset of successful trajectories. Let $\traj = (\obs_0, \action_0, \ldots, \obs_{T})$ denote a successful trajectory of length $T \leq H$. For each observation $\obs_t \in \traj$, we label its task progress using the normalized time remaining in the trajectory:
\begin{equation}
\progress_t = 1 - \frac{t}{T},
\label{eq:progress}
\end{equation}
such that $\progress_t$ decreases from $1$ at the start to $0$ at the end.

We construct our probing dataset $\mathcal{D}_\text{probe}$ by passing all $N$ successful demonstrations through the VLA and extracting the model's activations (as in Sec. \ref{sec:bg}):
\begin{equation}
\mathcal{D}_\text{probe}
=
\{(\activ_t^{0,n}, \activ_t^{1,n}  \hdots, \activ_t^{L,n}, \progress_t^n)\}_{n=1,\ldots,N;\;t=0,\ldots,T^n}.
\label{eq:data}
\end{equation}
Our probe is trained via linear regression to approximately predict progress via that layer's activations, $$\hat{\progress}(\obs) = \probe(\activ^i),$$ where $\activ^i$ is the activation of the $i$th layer when observation $\obs$ is input to the VLA model.

\para{Runtime Monitoring via  Progress Probes}
We convert our progress probe into a runtime monitor by comparing its prediction to the expected progress at the current timestep. Let $\mu_T$ denote the mean successful trajectory length: $ \mu_T = \frac{1}{N}\sum_{n=1}^N T^n.$ During \textit{nominal, successful} execution of the policy, the expected progress at time $t$ is simply $\progress^\text{nom}_t = 1 - \frac{t}{\mu_T}.$
Thus, our monitor score from Sec.~\ref{sec:problem} is simply the deviation from the expected nominal task progress time:
\begin{equation}
\score(\traj_t) = \hat{\progress}(\obs_t) - \progress^\text{nom}_t.
\end{equation}
Since larger progress values correspond to earlier stages of the task, a positive score indicates that the policy's estimated progress lags behind the nominal trajectory; a negative score indicates that it is ahead. We do not penalize faster execution.
Instead, we predict failures when execution falls sufficiently behind its nominal progression, i.e., when $s(\tau_t) > \delta$.

We note that this runtime monitoring recipe is simple by construction. The probe can be trained using only successful task executions and, because the probe is evaluated on the activations that are already computed as part of VLA inference, it requires minimal additional runtime computation.

\providecommand{\cmark}{\ensuremath{\checkmark}}
\providecommand{\xmark}{\ensuremath{\times}}
\begin{table*}[t]
\centering
\begin{minipage}[b]{0.67\textwidth}
  \centering
\scriptsize
\renewcommand{\arraystretch}{0.6}
\setlength{\tabcolsep}{3pt}
\resizebox{\linewidth}{!}{%
\begin{tabular}{lccccccccc}
\toprule
& \multicolumn{4}{c}{\textbf{Supervision}}
& \textbf{Speed}
& \multicolumn{2}{c}{\textbf{All tasks}}
& \multicolumn{2}{c}{\textbf{Unseen tasks}} \\
\cmidrule(lr){2-5}\cmidrule(lr){6-6}\cmidrule(lr){7-8}\cmidrule(lr){9-10}
\textbf{Method} & \textbf{Input}
& \textbf{Output}
& \textbf{Act.} & \textbf{Labels}
& \textbf{Latency (ms)} $\downarrow$
& \textbf{Episode} $\uparrow$ & \textbf{State} $\uparrow$
& \textbf{Episode} $\uparrow$ & \textbf{State} $\uparrow$ \\
\midrule
\methodFailInput~\cite{xu2025can} & \cmark & \xmark & \xmark & \xmark & $7.3\times10^{0}$ & 0.881 & 0.857 & 0.667 & 0.656 \\
\methodRndOE~\cite{romer2025failure} & \cmark & \xmark & \xmark & \xmark & $2.5\times10^{-1}$ & 0.874 & 0.850 & 0.661 & 0.557 \\
\midrule
\methodACE~\cite{romer2025failure} & \xmark & \cmark & \xmark & \xmark & $1.1\times10^{0}$ & 0.793 & 0.790 & 0.843 & 0.800 \\
\methodSTAC~\cite{agia2024sentinel} & \xmark & \cmark & \xmark & \xmark & $4.0\times10^{-1}$ & 0.477 & 0.442 & 0.482 & 0.434 \\
\midrule
\methodSafeMLP~\cite{gu2025safe} & \xmark & \xmark & \cmark & \cmark & $1.2\times10^{-1}$ & 0.930 & 0.894 & 0.896 & 0.866 \\
\methodSafeLSTM~\cite{gu2025safe} & \xmark & \xmark & \cmark & \cmark & $1.8\times10^{-1}$ & 0.923 & 0.851 & 0.859 & 0.803 \\
\methodFailFeature~\cite{xu2025can} & \xmark & \xmark & \cmark & \xmark & $1.7\times10^{-1}$ & 0.904 & 0.888 & 0.634 & 0.571 \\
\midrule
\methodRobometerProgress~\cite{liang2026robometer} & \cmark & \xmark & \xmark & \cmark & $5.5\times10^{2}$ & 0.779 & 0.666 & \textemdash & \textemdash \\
\methodRobometerSuccess~\cite{liang2026robometer} & \cmark & \xmark & \xmark & \cmark & $5.5\times10^{2}$ & 0.514 & 0.316 & \textemdash & \textemdash \\
\methodRobometerFTProgress & \cmark & \xmark & \xmark & \cmark & $5.5\times10^{2}$ & 0.922 & 0.830 & \textemdash & \textemdash \\
\methodRobometerFTSuccess & \cmark & \xmark & \xmark & \cmark & $5.5\times10^{2}$& 0.505 & 0.407 & \textemdash & \textemdash \\
\midrule
\methodOurs & \xmark & \xmark & \cmark & \xmark & $2.6\times10^{-2}$ & 0.909 & 0.788 & 0.900 & 0.794 \\
\bottomrule
\end{tabular}%
}

  \caption{\textbf{Simulation: Runtime Failure Monitors.} Input, Output, Act., and Labels indicate whether a method uses image-level inputs, output action distributions, internal VLA activations, and additional failure labels for training the runtime monitor, respectively. Latency is measured on an NVIDIA RTX A6000 GPU; Robometer uses a 10-frame context. Episode and State report AUROC for episode-level and state-level failure monitoring.}
  \label{tab:sim-monitors}
\end{minipage}\hfill
\begin{minipage}[b]{0.32\textwidth}
  \centering
  \includegraphics[width=\linewidth]{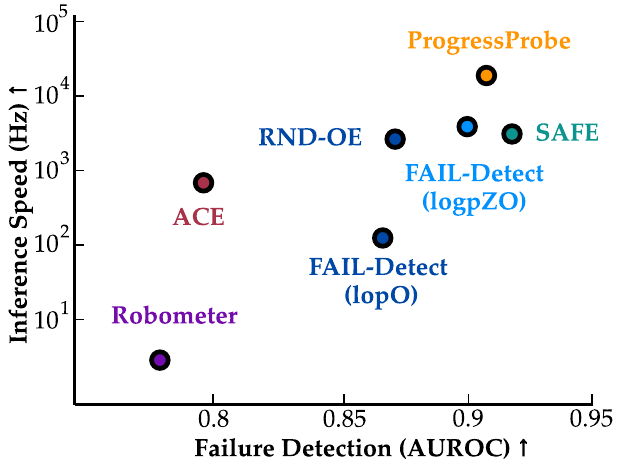}
  \captionsetup{type=figure}
  \vspace{-0.2in}
  \caption{\textbf{Accuracy vs. Speed.} The x-axis shows failure-detection AUROC, while the y-axis shows inference speed. Compared with prior runtime monitors, Progress Probe achieves a favorable balance of high AUROC and fast inference without relying on failure labels.}
  \label{fig:sim-runtime-summary}
  \vspace{0pt}
\end{minipage}
\vspace{-0.15in}
\end{table*}

\begin{table}[!t]
\centering
\footnotesize
\setlength{\tabcolsep}{4pt}
\resizebox{\columnwidth}{!}{%
\begin{tabular}{lcccc|c}
\toprule
& \multicolumn{4}{c}{TPR@5\%FPR by episode fraction $\uparrow$} & Det.\ time $\downarrow$ \\
\cmidrule(lr){2-5}
Detector & 25\% & 50\% & 75\% & 100\% & (frac.\ of ep.) \\
\midrule
\methodOurs & 0.00 & 0.19 & 0.54 & 0.64 & 0.60 \\
\methodSafeMLP & 0.00 & 0.08 & 0.41 & \textbf{0.77} & 0.75 \\
\methodSafeLSTM & \textbf{0.11} & \textbf{0.37} & 0.59 & 0.72 & 0.53 \\
\methodRobometerProgress & 0.01 & 0.08 & 0.21 & 0.33 & 0.67 \\
\methodRobometerFTProgress & 0.00 & 0.34 & \textbf{0.62} & 0.72 & 0.56 \\
\bottomrule
\end{tabular}}
\caption{\textbf{Simulation: Detection Time Results.}}
\label{tab:sim-rm-lead-time}
\vspace{-0.1in}
\end{table}

\section{Experiments: Simulation}
\label{sec:results-sim}
We first conduct simulation studies to understand (a) how discriminative and efficient progress probes are relative to other runtime monitors, (b) how progress probes compare to VLMs as runtime monitors, and (c) how progress probes of VLA internals, whose features have been learned using robot actions, compare to probes over VLM features, which are trained only on images and text.

\para{Setup} We study the $\pi_{0.5}$~\citep{intelligence2025pi05} VLA, fine-tuned on a dataset of $5,000$ trajectories ($500$ per manipulation task for $10$ tasks). All training and evaluations are done in VLABench~\citep{zhang2025vlabench}. For runtime monitoring, we evaluate on $6$ tasks for which the VLA achieves a non-negligible success rate.

\para{Training} We construct the probe dataset $\mathcal{D}_\text{probe}$ by passing $N=300$ trajectories ($50$ per task across the $6$ tasks) through the VLA and recording its activations, as in Eq.~\ref{eq:data}. We use an $80/20$ train-validation split of $\mathcal{D}_\text{probe}$ to train the probe, and use the $20\%$ validation split to calibrate the detection threshold following~\cite{romer2025failure}.

\para{Evaluation} We evaluate on $600$ VLA rollout trajectories ($100$ per task), evenly split between successful and failed executions. For each failed rollout, we manually label the timestep at which the failure occurs and mark all subsequent timesteps as failures for frame-level evaluation.

\para{Baselines} We compare Progress Probe ($\ours$) against prior runtime monitoring methods for VLAs, grouping them into three categories: \textbf{input}-level monitors that operate on visual observations, \textbf{feature}-level monitors that use internal VLA activations, and \textbf{output}-level monitors that operate on predicted actions. Table~\ref{tab:sim-monitors} summarizes the compared methods.

For \textbf{input}-level monitoring, we evaluate FIPER (RND-OE)~\citep{romer2025failure,burda2019exploration} and FAIL-Detect (logPO)~\citep{xu2025can}. We implement FAIL-Detect (logPO) using DINO image features rather than policy features to isolate monitoring based on the input observation. FIPER (RND-OE) corresponds to the observation-based component of FIPER and measures novelty in the visual input. For \textbf{feature}-level monitoring, we evaluate FAIL-Detect (logpZ0)~\cite{xu2025can}, SAFE-MLP, SAFE-LSTM~\citep{gu2025safe}, and $\ours$, all operating on the same VLA activations. For \textbf{output}-level monitoring, we evaluate FIPER (ACE) and STAC~\citep{agia2024sentinel}, which monitor the predicted action distribution. All prior monitors are unsupervised except SAFE, which uses failure labels to learn and calibrate a failure score.

We additionally compare against Robometer~\cite{liang2026robometer}, a VLM-based baseline trained on large-scale data containing both successful and failed trajectories to estimate task progress. This provides a direct comparison to $\ours$, which also estimates task progress but learns the progress signal from successful trajectories alone. SAFE is trained on these $300$ successful trajectories together with an additional $300$ failed trajectories for supervision and calibration, while Robometer is evaluated both zero-shot and after fine-tuning on the same success-and-failure dataset used for SAFE.

\subsection{On the Quality and Efficiency of Our Progress Probe}

We first evaluate progress probe (and all baselines) on three capabilities: (a) efficiency, (b) capability to discriminate success and failure, and (c) generalizability to unseen tasks.

\para{Setup} We break down our evaluation setup into:
\begin{enumerate}
\item \textbf{Efficency by Inference Latency}: All methods are run on an NVIDIA RTX A6000 for inference latency.
\item \textbf{Discriminability via Failure Prediction:} We evaluate runtime failure prediction on the $600$ VLA rollout trajectories at two levels: \textit{Episode} and \textit{State}. At the \textit{Episode} level, we evaluate whether the monitor discriminates successful from failed trajectories by taking the maximum monitor score over each trajectory. At the \textit{State} level, we evaluate whether the monitor discriminates successful from failed states using the score predicted at each timestep. We measure AUROC for failure prediction.
\item \textbf{Generalization to Unseen Tasks:} We evaluate how well failure discriminability generalizes to unseen tasks. For each split, we designate $3$ of the $6$ tasks as unseen and train the monitors on the remaining $3$ seen tasks, evaluating all ${6 \choose 3}=20$ possible splits. We exclude Robometer from this evaluation because it is primarily used zero-shot and is therefore not trained on the seen-task subset.
\item \textbf{Detection Time on Failures:} Following the objective in Eq.~\ref{eq:monitor-objective}, a good monitor should minimize detection time, $\DT(\delta,\traj;\score)$, while maintaining a low false-positive rate, $\FPR(\delta,\traj;\score)$. We therefore evaluate the \textit{average detection time} on failed rollouts. We calibrate the detection thresholds for $\ours$ and Robometer following \citep{romer2025failure}, while SAFE uses its prescribed conformal calibration procedure. At varying fractions of the episode, we report the true-positive rate (TPR) at a $5\%$ false-positive rate (FPR). We measure detection time as the fraction of the episode elapsed when the monitor first detects a failure, averaged across failed rollouts.
\end{enumerate}

\para{Result \#1: $\ours$ is Efficient Yet Effective} Fig.~\ref{fig:sim-runtime-summary} and Table~\ref{tab:sim-monitors} show that \textbf{feature}-level monitors are generally more efficient, as they directly leverage VLA activations already computed during policy inference. Among these methods, $\ours$ achieves the lowest inference latency of $2.6 \times 10^{-2}$ while maintaining competitive monitoring performance. By using a simple linear probe over VLA activations, $\ours$ is substantially faster than SAFE-MLP ($1.2 \times 10^{-1}$), SAFE-LSTM ($1.8 \times 10^{-1}$), and FAIL-Detect (logpZ0) ($1.7 \times 10^{-1}$), while achieving comparable failure-discrimination performance.

\para{Result \#2: $\ours$ Achieves Competitive Failure Discriminability} Overall, \textbf{feature}-level monitors achieve the strongest runtime monitoring performance, suggesting that internal VLA activations contain informative signals for distinguishing successful from failed executions. Notably, $\ours$ achieves an Episode AUROC of $0.909$ and a State AUROC of $0.788$ without using failure labels for training, remaining competitive with supervised baselines such as SAFE~\cite{gu2025safe}, which are trained with additional failure data.

\para{Result \#3: $\ours$ Generalizes Competitively to Unseen Tasks} $\ours$ remains competitive on unseen tasks, achieving an Episode AUROC of $0.900$ and a State AUROC of $0.794$. This is comparable to SAFE-MLP ($0.896$ Episode AUROC, $0.866$ State AUROC) and SAFE-LSTM ($0.859$ Episode AUROC, $0.803$ State AUROC). FIPER (ACE) also achieves similar performance ($0.843$ Episode AUROC, $0.800$ State AUROC), suggesting that both \textbf{feature}-level and \textbf{output}-level monitors can generalize effectively to new tasks. Overall, $\ours$ maintains competitive discriminability on unseen tasks while remaining more efficient than prior runtime monitors.

\para{Result \#4: $\ours$ Achieves Competitive Detection Time with High Accuracy} Table~\ref{tab:sim-rm-lead-time} shows that $\ours$ maintains competitive TPR throughout the episode, reaching $0.54$ at $75\%$ and $0.64$ at $100\%$ of the episode. On average, $\ours$ detects failures after $0.60$ of the episode, comparable to SAFE-LSTM at $0.53$. Notably, SAFE-LSTM explicitly models temporal history through its recurrent architecture, whereas $\ours$ uses only the current VLA activation with a simple linear probe. Despite this simplicity, $\ours$ achieves competitive detection timing, demonstrating that Progress Probe effectively captures the objective in Eq.~\ref{eq:monitor-objective} of detecting failures early while maintaining a low false-positive rate.

\begin{figure}[t]
\centering
\includegraphics[width=1.0\linewidth]{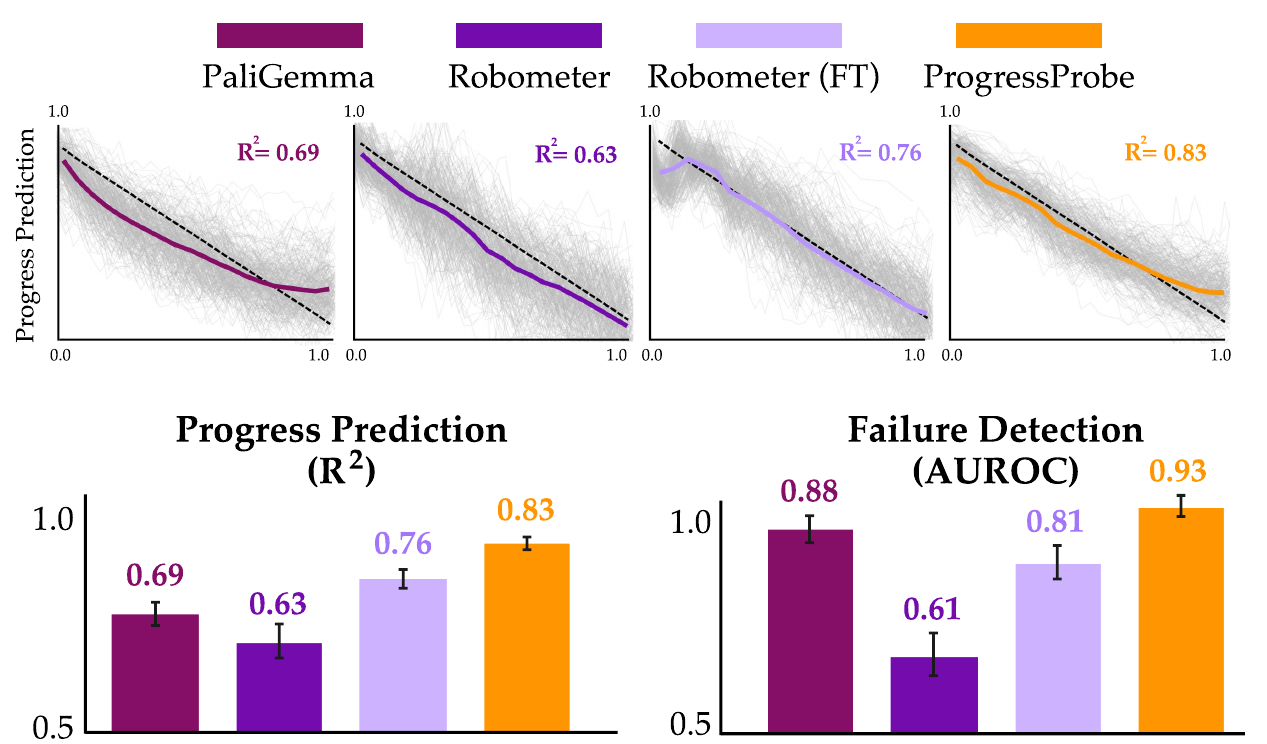}
\caption{\textbf{Simulation: Progress Probe vs. VLMs.} (\textit{Top}): Progress prediction results using progress probes trained on different representations. (\textit{Bottom}): Progress prediction mean-$R^2$ and failure detection results using progress probes trained on different representations.}
\label{fig:vla_vs_vlm}
\vspace{-0.15in}
\end{figure}

\subsection{VLA Progress Probes vs. VLMs as Runtime Monitors}

Next, we study how our lightweight progress probe compares to general-purpose VLMs as runtime monitors.

\para{Baselines} We compare Progress Probe ($\ours$) against the 4B-parameter Robometer~\citep{liang2026robometer}. We treat Robometer as a strong reference for task-progress estimation because it is trained on diverse robot data across hardware and simulation, including successful, failed, and intermediate-progress examples. Robometer provides two prediction heads: \textit{Progress}, which estimates the current task progress, and \textit{Success}, which estimates the probability of task success and is primarily intended as a terminal signal. We evaluate both heads within the same runtime monitoring framework defined in Sec.~\ref{sec:methods}.

To calibrate the Progress head, we run Robometer on the same $300$ training trajectories used for Progress Probe and aggregate its predicted progress curves for each task, defining nominal progress from Robometer's own predictions on successful trajectories. We evaluate two variants: Robometer-ZS (zero-shot) and Robometer-FT, fine-tuned on the same success-and-failure dataset used to train SAFE.

\para{Results: Progress Probes are Fast \& Competitive with VLMs}
Fig.~\ref{fig:sim-runtime-summary} and Table~\ref{tab:sim-monitors} show that $\ours$ and Robometer-FT achieve comparable success-failure discriminability. However, $\ours$ fine-tunes only $2$K parameters compared with $4$B parameters for Robometer-FT and runs over $10{,}000\times$ faster at inference time. Thus, $\ours$ achieves performance comparable to a large VLM while being substantially more compact and computationally efficient.

\subsection{Does Robot Action Data Matter for Progress Probing?}
Unlike VLMs, which are trained only on images and text, VLAs are additionally fine-tuned on robot \textit{action} data. We ask whether this action-grounded training yields internal activations that better encode task progress by comparing the performance of progress probes trained on VLA activations and those trained on VLM activations.

\para{Methods} We compare our progress probe trained on VLA activations ($\ours_{\pi_{0.5}}$) to PaliGemma-Probe ($\ours_\text{Pali}$, Robometer-ZS-Probe ($\ours_\text{Robo-ZS}$), and Robometer-FT-Probe ($\ours_\text{Robo-FT}$). The latter are trained on the corresponding VLM's activations in the same process we trained the original progress probe. We use the 3B PaliGemma which is the backbone of $\pi_{0.5}$.

\para{Metrics} We measure the progress prediction from the probe using $R^2$ on the success rollouts from the evaluation data split. We also measure the success/failure AUROC of the probes as in the previous experiment.

\para{Results: Robot Action Data Improves Progress Probe Performance}
As seen in Fig.~\ref{fig:vla_vs_vlm}, $\ours_{\pi_{0.5}}$ performs the best at predicting progress on successful rollouts with an $R^2$ of $0.83$ compared to $\ours_\text{Robo-ZS}$ ($R^2 = 0.63$), $\ours_\text{Robo-FT}$ ($R^2 = 0.76$), and $\ours_\text{Pali}$ ($R^2 = 0.69$).
We find that models that see \textit{visuals} of robots in training (like in Robometer) do not lead to improvement over a standard VLM  when it comes to predicting progress. Additionally, when it comes to monitoring, $\ours_{\pi_{0.5}}$ achieves an AUROC of $0.93$ on the simulation evaluation set, outperforming of $\ours_\text{Robo-ZS}$ (AUROC: 0.61), $\ours_\text{Robo-FT}$ (AUROC: 0.81), and $\ours_\text{Pali}$ (AUROC: 0.88). We hypothesize this difference is caused by the VLA training objective, i.e., learning to predict actions has caused the policy to develop internal representations that can better detect when the policy is entering failure. We argue this result points to the importance of not just robot data, but action data, in understanding whether the policy will succeed or fail in its current rollout.

\begin{table}[t]
\centering

\scriptsize
\renewcommand{\arraystretch}{1.0}
\setlength{\tabcolsep}{3pt}
\resizebox{\columnwidth}{!}{%
\begin{tabular}{l*{7}{c}}
\toprule
\textbf{Method} & \shortstack{\textbf{Block}\\\textbf{Stack}} & \shortstack{\textbf{Knife}\\\textbf{Put}} & \shortstack{\textbf{Stacked}\\\textbf{Utensil}} & \shortstack{\textbf{Candy}\\\textbf{Bag}} & \shortstack{\textbf{Coffee}\\\textbf{Bag}} & \shortstack{\textbf{Coffee}\\\textbf{Bean}} & \shortstack{\textbf{Average}\\\textbf{AUROC}} \\
\midrule
\methodFailInput & 0.40 & 0.46 & 0.59 & 0.67 & 0.49 & 0.42 & 0.54 \\
\methodRndOE & 0.61 & 0.82 & 0.74 & 0.50 & 0.60 & 0.52 & 0.61 \\
\midrule
\methodACE & 0.61 & 0.68 & 0.62 & 0.50 & 0.60 & 0.50 & 0.57 \\
\methodSTAC & 0.59 & 0.57 & 0.37 & 0.31 & 0.57 & 0.34 & 0.45 \\
\midrule
\methodSafeMLP & 0.60 & 0.80 & 0.60 & 0.72 & 0.39 & 0.38 & 0.73 \\
\methodSafeLSTM & 0.70 & 0.70 & 0.57 & 0.76 & 0.50 & 0.33 & 0.58 \\
\methodFailFeature & 0.47 & 0.94 & 0.79 & 0.90 & 0.72 & 0.74 & 0.69 \\
\midrule
\methodRobometerProgress & 0.53 & 0.51 & 0.73 & 0.94 & 0.61 & 0.66 & 0.70 \\
\methodRobometerSuccess & 0.33 & 0.33 & 0.30 & 0.46 & 0.24 & 0.25 & 0.41 \\
\midrule
\methodRobometerFTProgress & 0.63 & 0.56 & 0.88 & 0.95 & 0.66 & 0.53 & 0.75 \\
\methodRobometerFTSuccess & 0.32 & 0.43 & 0.28 & 0.39 & 0.20 & 0.18 & 0.34 \\
\midrule
\methodOurs & 0.87 & 0.80 & 0.81 & 0.92 & 0.66 & 0.26 & 0.78 \\
\bottomrule
\end{tabular}%
}
\caption{\textbf{Hardware: All-task.} State-level failure-detection AUROC is reported for each task. The rightmost column reports the average performance across all six tasks.}
\label{tab:real-all-per-task}
\end{table}

\begin{table}[t]
\centering
\scriptsize
\renewcommand{\arraystretch}{1.0}
\setlength{\tabcolsep}{3pt}
\resizebox{\columnwidth}{!}{%
\begin{tabular}{l*{7}{c}}
\toprule
\textbf{Method} & \shortstack{\textbf{Block}\\\textbf{Stack}} & \shortstack{\textbf{Knife}\\\textbf{Put}} & \shortstack{\textbf{Stacked}\\\textbf{Utensil}} & \shortstack{\textbf{Candy}\\\textbf{Bag}} & \shortstack{\textbf{Coffee}\\\textbf{Bag}} & \shortstack{\textbf{Coffee}\\\textbf{Bean}} & \shortstack{\textbf{Average}\\\textbf{AUROC}} \\
\midrule
\methodFailInput & 0.51 & 0.66 & 0.54 & 0.55 & 0.60 & 0.49 & 0.56 \\
\methodRndOE & 0.40 & 0.83 & 0.76 & 0.62 & 0.51 & 0.56 & 0.61 \\
\midrule
\methodACE & 0.61 & 0.68 & 0.62 & 0.50 & 0.60 & 0.50 & 0.59 \\
\methodSTAC & 0.59 & 0.57 & 0.37 & 0.31 & 0.57 & 0.34 & 0.46 \\
\midrule
\methodSafeMLP & 0.60 & 0.78 & 0.64 & 0.75 & 0.53 & 0.31 & 0.60 \\
\methodSafeLSTM & 0.54 & 0.64 & 0.56 & 0.70 & 0.51 & 0.14 & 0.51 \\
\methodFailFeature & 0.48 & 0.80 & 0.77 & 0.83 & 0.50 & 0.75 & 0.69 \\
\midrule
\methodOurs & 0.87 & 0.75 & 0.84 & 0.94 & 0.60 & 0.84 & 0.81 \\
\bottomrule
\end{tabular}%
}\caption{\textbf{Hardware: Unseen Task.} State-level failure detection AUROC is reported for each task. The rightmost column reports the average performance across all six tasks.}
\label{tab:real-loto-per-task-unseen}
\vspace{-0.2in}
\end{table}

\section{Experiments: Hardware}
\label{sec:results-hardware}

Finally, we study progress probes on hardware with a suite of manipulation tasks of increasing physical complexity.

\para{Setup} Our hardware setup is a Franka Research 3 robotic arm with the DROID \cite{khazatsky2024droid} camera setup. Our VLA is $\pi_{0.5}$~\citep{intelligence2025pi05}. We use the $\pi_\text{0.5-DROID}$ checkpoint and finetune it on our $6$ hardware tasks with $60$ trajectories per task ($360$ trajectories total), from StressDream~\citep{seo2026stressdream}. Our hardware tasks are:
\begin{enumerate}
    \item \textbf{Candy Bag}: Pick and place an open candy bag into a tan bin without spilling the candy.
    \item \textbf{Coffee Bag}: Pick and place an open coffee bag into a tan bin without spilling the coffee beans.
    \item \textbf{Coffee Bean}: Pour coffee beans into the bowl without spilling the coffee beans.
    \item \textbf{Knife Put}: Pick and place the knife into a bowl.
    \item \textbf{Block Stack}: Pick and place the orange Jenga block onto the blue Jenga block.
    \item \textbf{Stacked Utensil}: Pick up the blue spoon without pushing the brown spoon off the plate.
\end{enumerate}

\paras{Training}
Our probe training data is obtained by passing $360$ hardware trajectories (from the VLA fine-tuning dataset) through the policy to get corresponding activation data. We label the data with the progress labels as in Eq.~\ref{eq:progress}.
The probe is trained on $80/20$ splits for training and validation.

\paras{Evaluation} We evaluate all methods using 120 rollouts from the VLA policy (20 rollouts per task) with a 50/50 split for success and failure. For all failure rollouts, we manually label where the failure occurs in the episode and mark the remainder of the trajectory as a failure for frame-level evaluation.

\begin{figure}[t]
\centering
\includegraphics[width=0.85\linewidth]{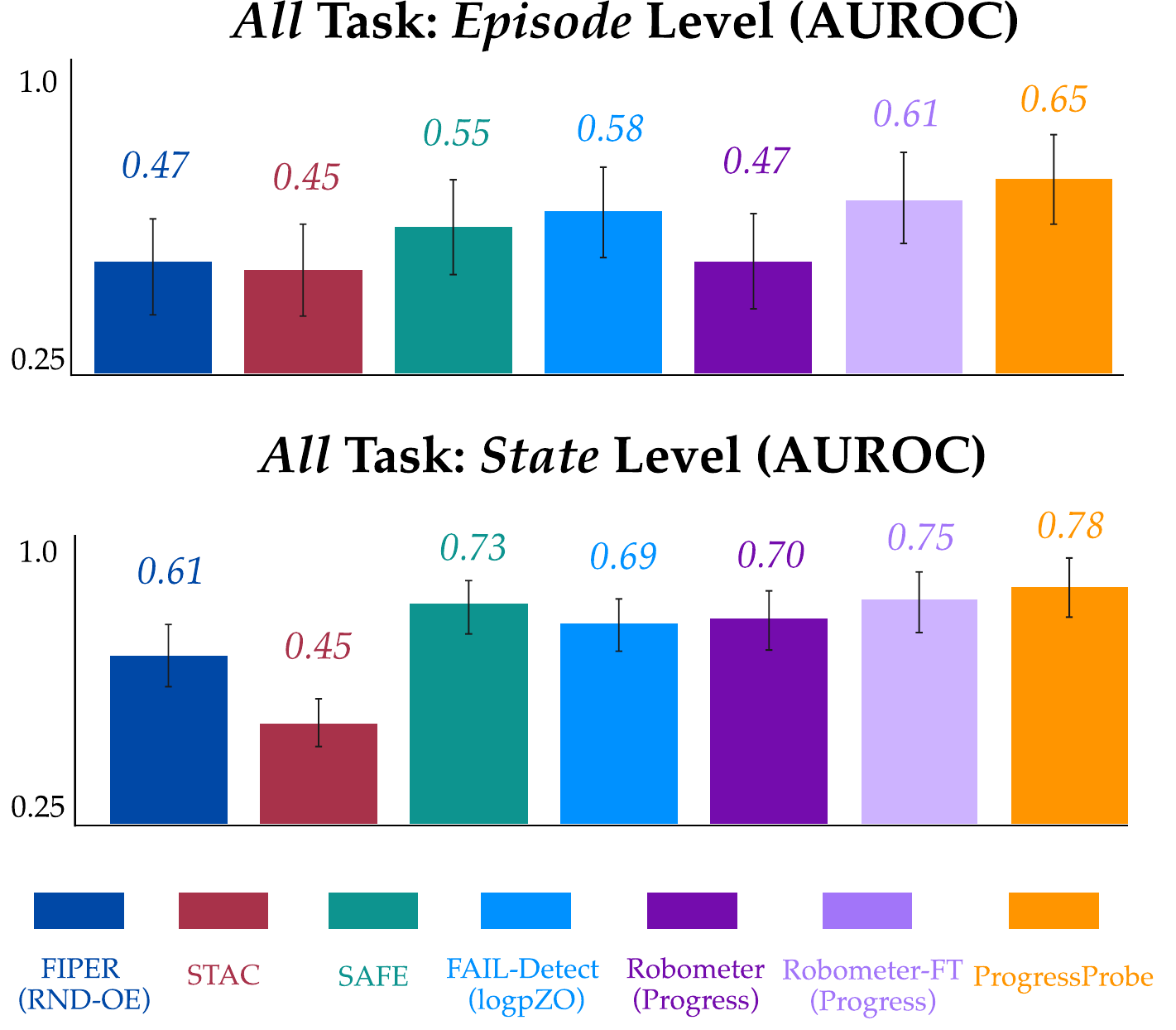}
\caption{\textbf{Hardware: Runtime Monitoring.} Episode-level and state-level AUROC for failure detection. $\ours$ achieves strong discriminability at both levels for runtime monitoring.}
\label{fig:hw_quantitative}
\vspace{-0.15in}
\end{figure}

\para{Baselines} We compare Progress Probe ($\ours$) against the same runtime monitoring baselines used in simulation, adopting the same implementations and monitor categories: \textbf{input}-level methods (FIPER (RND-OE)~\citep{romer2025failure,burda2019exploration} and FAIL-Detect (logPO)~\citep{xu2025can}), \textbf{feature}-level methods (FAIL-Detect (logpZ0)~\cite{xu2025can}, SAFE-MLP, and SAFE-LSTM~\citep{gu2025safe}), and \textbf{output}-level methods (FIPER (ACE) and STAC~\citep{agia2024sentinel}). We additionally compare against Robometer~\cite{liang2026robometer} in both zero-shot and fine-tuned variants. As in simulation, SAFE and Robometer-FT use additional failure data, while the remaining monitors, including $\ours$, are trained using successful trajectories only.

\subsection{Discriminability by Failure Detection}
We test if $\ours$ scales to monitoring a real-world VLA policy, evaluating how well $\ours$ distinguishes successful from failed execution when deployed on the real robot.

\para{Method} We evaluate each baseline on the hardware evaluation dataset and ask the monitor to determine (a) if the entire episode is a successful or failed rollout via max aggregation and (b) given a trajectory, if the current time-step is a successful state or failure state.

\begin{figure}[t]
\centering
\includegraphics[width=1.0\linewidth]{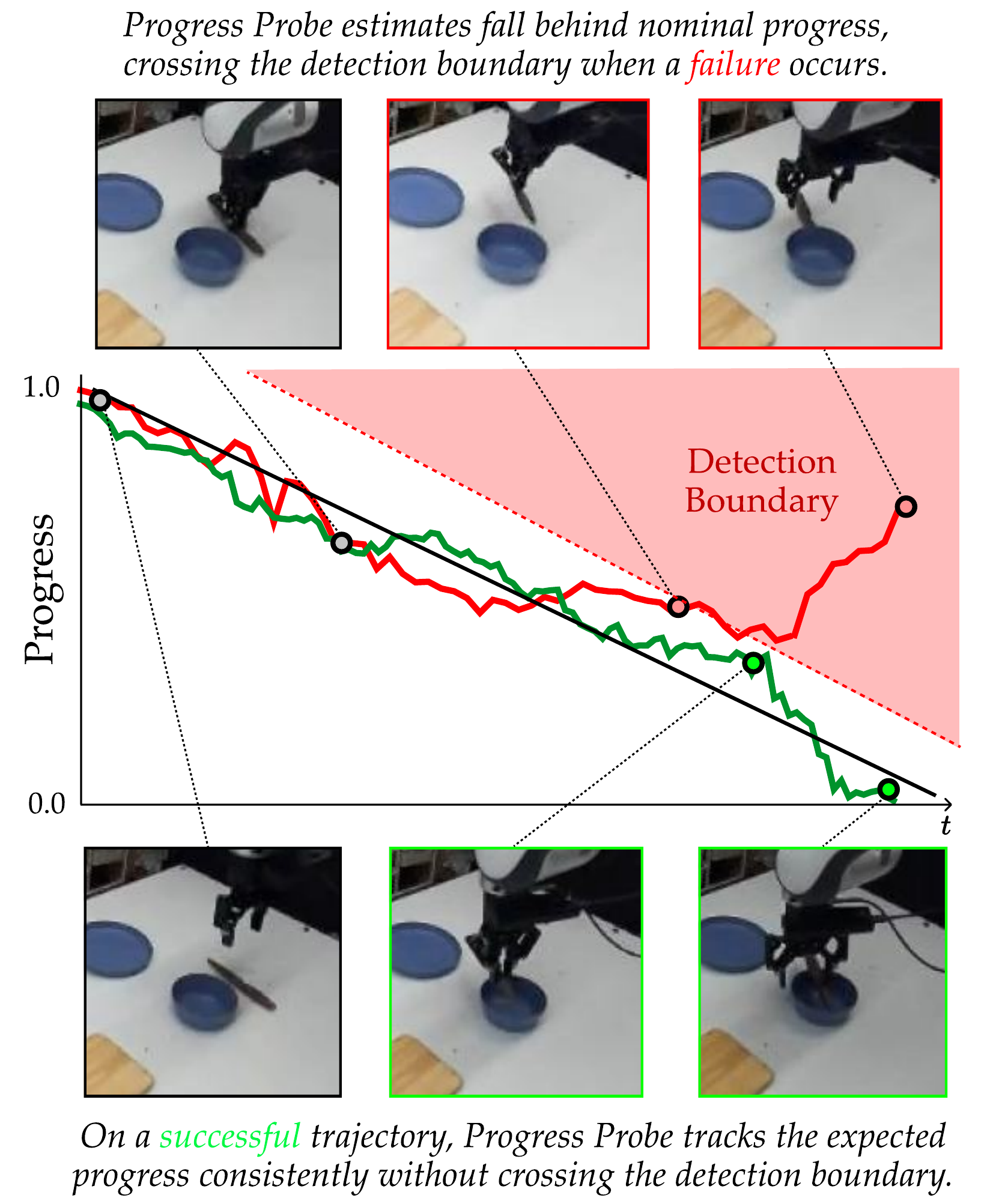}
\caption{\textbf{Hardware: Qualitative.} Red curves show failed trajectories and $\ours$ predictions, while green curves show predictions on successful trajectories for the Knife Put task. $\ours$ correctly identifies failures while remaining within the nominal range on successful executions, demonstrating reliable separation between successful and failed behavior.}
\label{fig:hw_qualitative}
\vspace{-0.15in}
\end{figure}

\para{Result: $\ours$ Better Discriminates Success and Failure at Both Episode and State Levels} Fig.~\ref{fig:hw_quantitative} shows that $\ours$ achieves higher Episode- and State-level AUROC than all baselines, including Robometer-FT and SAFE, which require additional failure trajectories and labels for training. Fig.~\ref{fig:hw_qualitative} further shows that $\ours$ qualitatively separates successful and failed executions: predicted progress falls behind the nominal trajectory and crosses the detection boundary near the onset of failure, while remaining within the nominal range on successful rollouts.

Table~\ref{tab:real-all-per-task} further shows that $\ours$ performs strongly across hardware tasks, achieving the highest AUROC on Block Stack ($0.87$) and Stacked Utensil ($0.81$), while remaining competitive on the remaining tasks. $\ours$ underperforms on more challenging tasks such as Coffee Bean, where failures can arise stochastically despite in-distribution actions and may be visually subtle due to the small image region associated with the failure. We hypothesize that, in such cases, the VLA activations may not change sufficiently to provide a strong failure signal. Mirroring the simulation results, \textbf{feature}-level monitors generally perform strongly on hardware, while fine-tuned Robometer achieves comparable performance on several tasks. The task-dependent variation across methods suggests that internal VLA features and visual task-progress cues provide complementary signals for runtime failure monitoring.

\subsection{How Well Do $\ours$ Monitors Generalize to Unseen Tasks?}

Next, we test the generalizability of progress probes and baseline runtime monitors to unseen tasks in hardware.

\para{Method} Given that we have $6$ tasks, we choose $3$ tasks randomly to be unseen and repeat till all ${6 \choose 3} = 20$ combinations are exhausted. We re-train our method and all baselines on the $3$ seen tasks, and evaluate on the unseen tasks' rollouts from the evaluation set at the State level. We do not include Robometer since this VLM is typically used zero-shot, so all tasks would be ``unseen'' at all times.

\para{Result: $\ours$ Generalizes Better to Unseen Tasks} Table~\ref{tab:real-loto-per-task-unseen} shows that $\ours$ generalizes strongly to held-out tasks, achieving the highest State AUROC of $0.81$ averaged across all ${6 \choose 3}=20$ seen/unseen task splits. We hypothesize that this generalization arises from probing internal VLA activations, which encode shared representations learned across diverse tasks and hardware interactions. Per-task results further show that $\ours$ consistently generalizes well across the held-out tasks, suggesting that Progress Probe captures task-progress structure shared across diverse manipulation settings.

\section{Conclusion and Limitations}
\label{sec:discussion}

In this work, we introduced \textit{Progress Probes} ($\ours$), a lightweight approach to runtime monitoring that linearly decodes task progress from a VLA's internal activations. Across simulation and hardware experiments, $\ours$ achieves competitive failure-monitoring performance while being substantially more computationally efficient than existing VLM-based monitors. These results suggest that task progress encoded within VLA activations provides a simple yet effective signal for runtime failure monitoring.

\para{Limitations} Despite promising results across simulation and hardware, several limitations remain. First, our experiments focus on $\pi_{0.5}$~\cite{intelligence2025pi05}, and future work should investigate how Progress Probes transfer across different VLA families, such as GR00T~\cite{nvidia2025gr00t} and MolmoAct~\cite{fang2026molmoact2}, whose architectures and training recipes induce different internal representations. Second, we primarily probe activations from the VLM backbone of the VLA. Many recent VLAs additionally employ continuous action experts, such as diffusion or flow-matching models; investigating whether their internal activations provide complementary signals for runtime monitoring is an important direction for future work.

\section*{Acknowledgments}
This research was supported in part by funding from the Defense Advanced Research Projects Agency (DARPA) Young Faculty Award (YFA) and Schmidt Sciences. The views, opinions and/or findings expressed are those of the author and should not be interpreted as representing the official views or policies of DARPA or the U.S. Government. AB was supported by the National Science Foundation Graduate Research Fellowship Program under Grant No. DGE2631988.

\renewcommand{\bibfont}{\footnotesize}
\bibliographystyle{IEEEtranN}
\bibliography{references}

\end{document}